\documentclass[sigconf]{acmart}
\usepackage{hyperref}
\AtBeginDocument{%
  \providecommand\BibTeX{{%
    \normalfont B\kern-0.5em{\scshape i\kern-0.25em b}\kern-0.8em\TeX}}}

\copyrightyear{2024} 
\acmYear{2024} 
\setcopyright{acmlicensed}\acmConference[COMPASS '24]{ACM SIGCAS/SIGCHI Conference on Computing and Sustainable Societies}{July 8--11, 2024}{New Delhi, India}
\acmBooktitle{ACM SIGCAS/SIGCHI Conference on Computing and Sustainable Societies (COMPASS '24), July 8--11, 2024, New Delhi, India}
\acmDOI{10.1145/3674829.3675076}
\acmISBN{979-8-4007-1048-3/24/07}

\begin{document}

\title{Designing A Sustainable Marine Debris Clean-up Framework without Human Labels}


\author{Raymond Wang}
\email{ray.wang@colby.edu}
\affiliation{%
  \institution{Davis Institute for AI, Colby College}
  \city{Waterville}
  \state{Maine}
  \country{USA}
}

\author{Nicholas R. Record}
\email{nrecord@bigelow.org}
\affiliation{%
  \institution{Bigelow Laboratory for Ocean Sciences}
  \city{East Boothbay}
  \state{Maine}
  \country{USA}
}

\author{D. Whitney King}
\email{dwking@colby.edu}
\affiliation{%
  \institution{Department of Chemistry, Colby College}
  \city{Waterville}
  \state{Maine}
  \country{USA}
}

\author{Tahiya Chowdhury}
\email{tahiya.chowdhury@colby.edu}
\affiliation{%
  \institution{Davis Institute for AI, Colby College}
  \city{Waterville}
  \state{Maine}
  \country{USA}
}


\renewcommand{\shortauthors}{Wang et al.}

\begin{abstract}


Marine debris poses a significant ecological threat to birds, fish, and other animal life. Traditional methods for assessing debris accumulation involve labor-intensive and costly manual surveys. This study introduces a framework that utilizes aerial imagery captured by drones to conduct remote trash surveys. Leveraging computer vision techniques, our approach detects, classifies, and maps marine debris distributions. The framework uses Grounding DINO, a transformer-based zero-shot object detector, and CLIP, a vision-language model for zero-shot object classification, enabling the detection and classification of debris objects based on material type without the need for training labels. To mitigate over-counting due to different views of the same object, Scale-Invariant Feature Transform (SIFT) is employed for duplicate matching using local object features. Additionally, we have developed a user-friendly web application that facilitates end-to-end analysis of drone images, including object detection, classification, and visualization on a map to support cleanup efforts. Our method achieves competitive performance in detection (0.69 mean IoU) and classification (0.74 F1 score) across seven debris object classes without labeled data, comparable to state-of-the-art supervised methods. This framework has the potential to streamline automated trash sampling surveys, fostering efficient and sustainable community-led cleanup initiatives. 

\end{abstract}



\maketitle

\section{Introduction}

The oceans, covering 71\% of the earth's surface, serve as a vast repository of biodiversity and sustain marine life. This precious ecosystem is threatened as various human-made materials, known as `marine debris', enter the ocean. Approximately 75 to 199 million tons of plastic is currently found in our oceans, resulting in severe environmental, social, economic, and health consequences, according to a report by the United Nations Environment Programme~\cite{UNEPreport}. The accumulation of these marine debris objects in a specific region is exacerbated by ocean currents and weather patterns, often linking the marine debris problem to climate change trends for particular oceanic regions~\cite{LINCOLN2022155709}. Marine debris includes a spectrum of persistent solid materials such as metals, plastics, rubber, fishing gear, and other discarded items that are manufactured or processed, and abandoned in the marine environment. Government, industry, and community stakeholders are developing marine debris action plans tailored to the local community's needs, which involve monitoring, coordination, collection, and removal efforts. However, implementing such initiatives can be expensive, both in terms of funding and human efforts.

\begin{figure}[t!] 
  \centering 
  \includegraphics[width=0.46\textwidth]{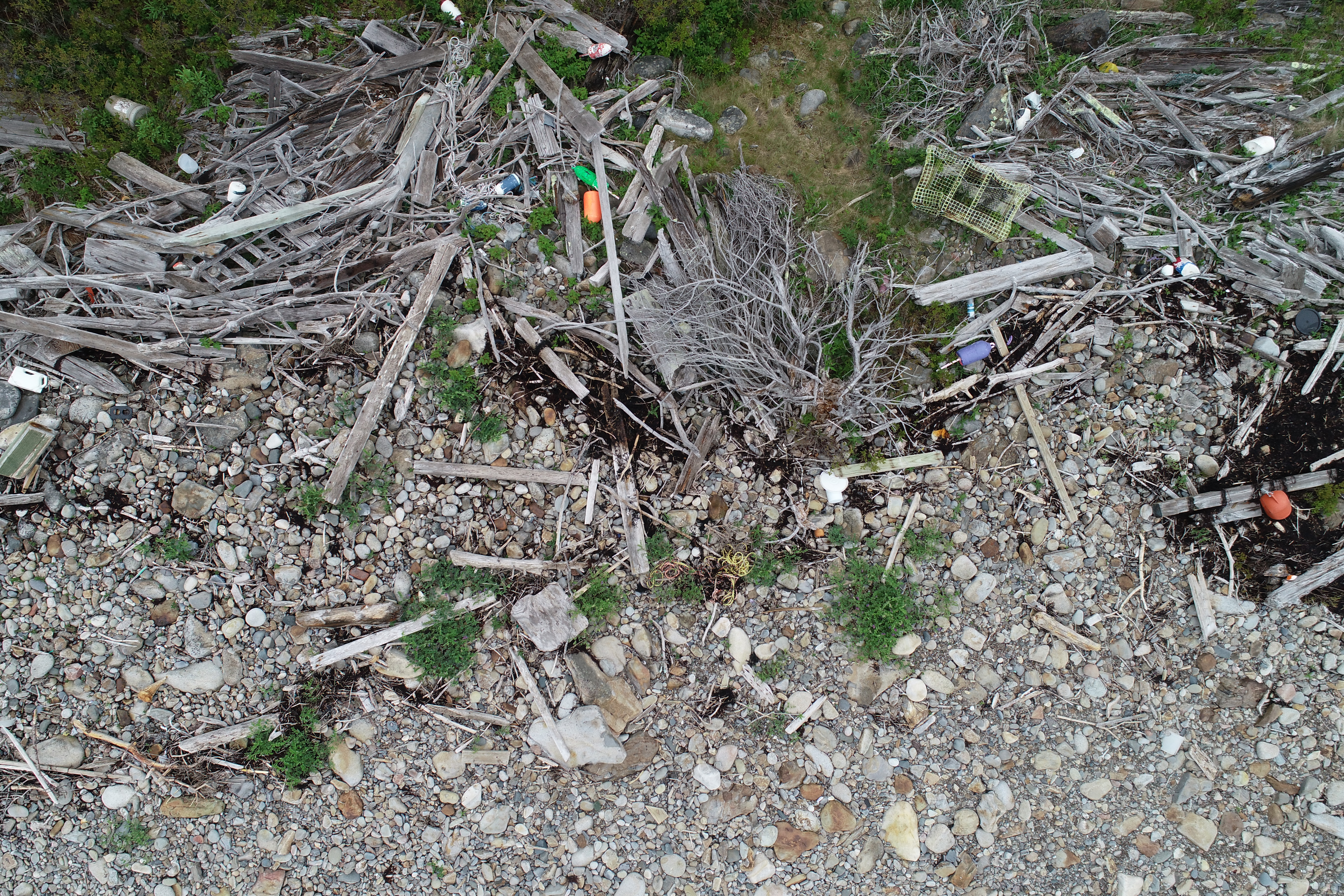}
  \caption{Beverage bottle, fishing gear and other marine debris along a beach on Allen Island in Maine, USA where the data was collected.}
  \Description{An aerial image with beverage bottle, fishing gear, and other marine debris found along a beach on Allen Island in Maine, USA where the data was collected.}
  \label{fig:top_example}
\end{figure}

To accelerate monitoring and clean-up efforts, previous research has developed remote sensing and computer vision-based approaches to detect and classify marine debris objects. Multi-spectral measurements available from image sensors have been used to develop spectral indices to distinguish plastic and other processed materials from natural materials such as vegetation and water~\cite{essd-12-2665-2020, essd-13-713-2021, tasseron2021advancing, freitas2021remote}. With recent advancements of deep learning in computer vision tasks, prior research has used object segmentation and detection approaches for automating marine debris monitoring from aerial imagery captured by satellite and drones~\cite{martin2021enabling, iordache2022targeting, pfeiffer2022use}. However, identifying marine debris objects from an image taken from the top view and classifying it based on appropriate material type still remains challenging~\cite{WINANS2023103515}. Most debris objects are often obscured by natural objects in the background (rocks, grass, bushes, wetlands), making it difficult to distinguish debris objects from the background land cover (see Figure~\ref{fig:top_example}). Additionally, marine debris objects can be highly diverse in shape, size, color, and texture which is dependent on economic, industry, and marine activities of the specific region. To resolve this challenge, human interpretation is typically leveraged as a sustainable approach to detect and identify debris items to monitor an area without making a physical visit~\cite{moy2018mapping, merlino2021citizen}.

\begin{figure*}[t!] 
  \centering 
  \includegraphics[width=0.90\textwidth]{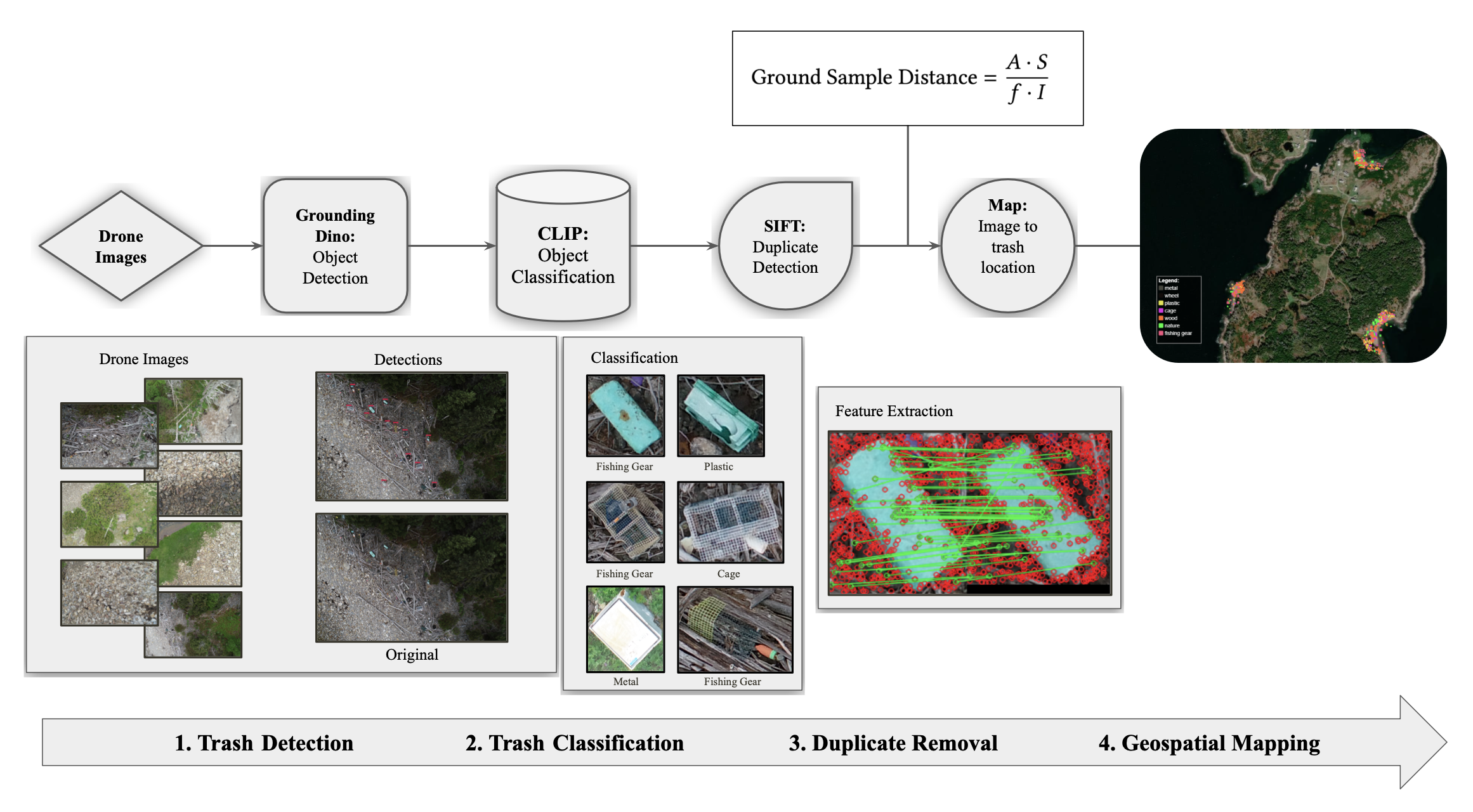}
  \caption{System diagram of the proposed framework.}
  \Description{System diagram of the proposed framework with four steps: trash detection, classification, duplicate removal and map visualization.}
  \label{fig:Implementation}
\end{figure*}

Typical machine learning-based solutions rely on human-provided decisions about the location and material type as ground truth to train algorithms to identify similar items from images collected in the future. However, human annotation is a time-consuming, tedious process that can take up to months, hence is costly to scale across large geographic regions~\cite{moy2018mapping}. Additionally, while relying on an expert annotator familiar with the geographical, environmental, and economic context is desirable, the effectiveness of such solutions remains limited to a specific dataset or geographic region. As a result, most solutions are designed around training custom ML-based frameworks that involve data pre-processing, annotation, model training, and validation procedures led by researchers and technical teams, which are often challenging to adopt by local communities. There is a need to design a sustainable, user-friendly framework for marine debris detection and cleanup that combines state-of-the-art computer vision tools and local stakeholders' knowledge.

To address this gap, this paper introduces a framework that detects, classifies, and maps the distribution of marine debris objects facilitating longitudinal monitoring and expedited clean-up efforts. As an alternative to cost-prohibitive ground-level sampling and time-consuming human inspection methods, our approach leverages existing computer vision models trained on large-scale image data to monitor marine debris accumulation over time. Specifically, we used Grounding DINO, an object detection model to query specific visual concepts related to custom marine debris objects in a coastal region such as islands and beaches without manual annotation. To streamline collection efforts, debris objects are categorized using an open-set image classifier that can be prompted to classify images with natural language debris categories. To improve the accessibility of the framework, we have designed an interactive web application that enables users to add aerial images, and analyze and map the spatial distribution of debris to co-ordinate the clean-up initiatives effectively. We evaluate the framework using a novel aerial image dataset captured via drone on a small coastal island, and we also discuss future directions for local and global scale initiatives. We have made the framework, web application, and dataset used in this study publicly available\footnote{\url{https://github.com/Tahiya31/Trash_Track}}.







\section{Method}

In this paper, we present a framework for detecting marine debris objects from drone images, classify them by object types, and visualize their location distribution on a map. A system diagram of different components of the framework is presented in Figure~\ref{fig:Implementation}. We describe the four key components of the framework in this section: object detection, classification, duplicate removal, and map visualization. 


\subsection{Detection using Grounding DINO}

\subsubsection{Zero-shot object detection} Existing works on marine debris or waste  detection formulate this problem as object detection, which is a computer vision task to identify certain objects in an image. The training procedure for such models often involves training supervised detection models using a collection of annotated images labeled by a human, called `training data'. Manual annotation is time-consuming and does not scale well for diverse geographic regions as marine debris categories can vary widely depending on social, economic, and geographic factors. Since our objective is to develop an automated pipeline with minimal annotation efforts, we relied on existing models trained on large-scale image datasets for object detection tasks in a zero-shot learning setting. Zero-shot object detector is trained on large image datasets with the intention that it can learn to recognize objects unseen during the training. Traditional object detection models require to be trained explicitly on examples of images containing object categories that it will need to detect in the evaluation phase, in a supervised learning setting. On the other hand, a zero-shot detector leverages its knowledge block obtained from pre-training on large-scale training data for open-set object detection without labels when encountered by novel categories and conditions.

To detect marine debris objects present in the drone images, we selected Grounding DINO, a zero-shot object detector~\cite{liu2023grounding}. It is designed for object detection in open set scenarios so that arbitrary objects can be detected based on human-provided inputs such as a category or a label. We select Grounding DINO for our purpose as it leverages cross-modal information from vision and language for open set concept generalization. As a result, Grounding DINO can generate object region proposals, indicating an object's location within an image, based on human-provided class labels, and detect novel categories.

\subsubsection{Parameters for Grounding DINO}
Our objective is to detect the objects relevant to marine debris from the images. Following the official implementation of Grounding DINO, we provided a set of image and text pairs as inputs to the model. The text input is used as a query command to detect all instances of that category in each image, which, in this case, was `all trash'. Since our dataset consists of drone images captured in coastal islands, the background landscape contained various natural objects such as woodland, water, rocks, etc. Our preliminary analysis showed that rocks were often categorized as trash due to lighting and angle, hence increasing false positives. To resolve this, we employed a second query for `all rocks'. Once the detected objects were retrieved, we filtered and removed the instances of trash that overlapped with where rocks were detected to lower the false positive rates resulting from the background landcover.

The Grounding DINO provides detected objects and their labels based on a user-provided threshold. Each detected box is accompanied by a similarity score to indicate its similarity to each of the text queries. The box threshold is used to prune the objects detected in an image (maximum limit is 900 objects) that fall below the threshold, which is analogous to a confidence score for an object. Similarly, a text threshold can be set so that final labels for a detected object are chosen based on the threshold. We experimented with different threshold levels as we observed that different levels were effective for different categories of debris objects. In our experimentation with the dataset, we empirically found 0.3 and 0.15 for both text and box thresholds to be effective as they provided detection performance comparable to human detection. We ran the object detection queries with these two threshold levels twice and combined the detection results to leverage their individual effectiveness.

\begin{figure}[t]
  \centering
  \includegraphics[width=\linewidth]{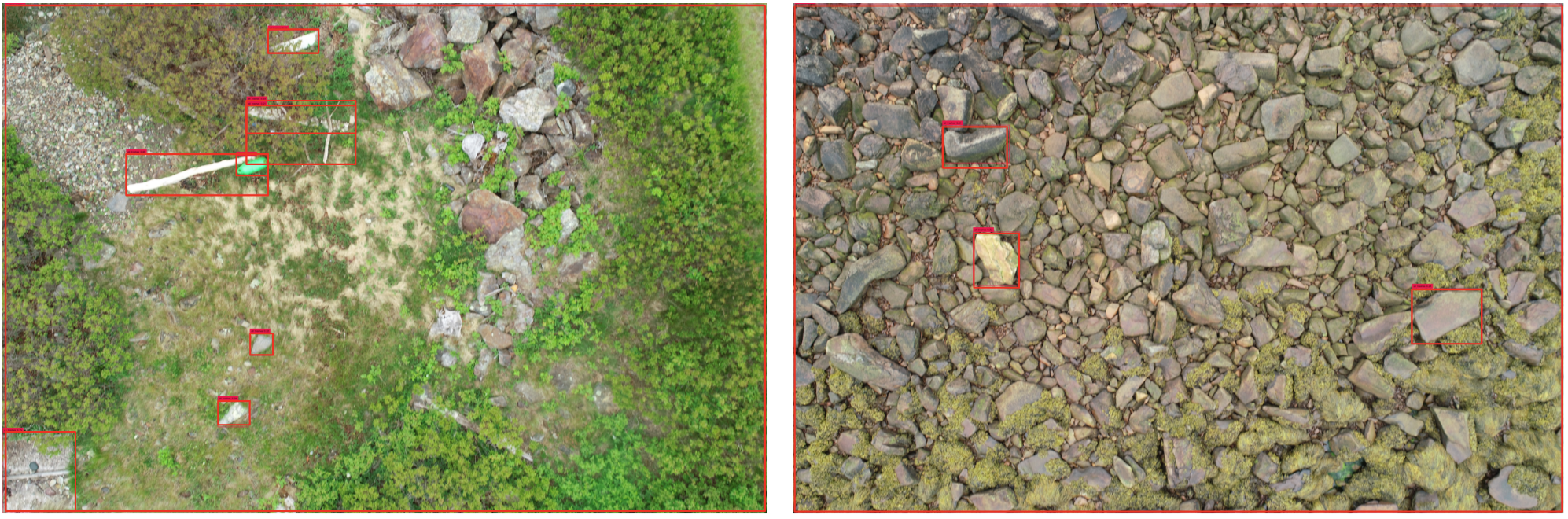}
  \caption{Examples of detections obtained from the raw query "all trashes." Detections are highlighted using red bounding boxes to indicate the identified areas within the dataset.}
  \label{fig: misclassification}
  \Description{Two example for misclassification where  detections obtained from the raw query "all trashes" are shown. Detections are highlighted using red bounding
boxes to indicate the identified areas.}
  \vspace{-6pt}
\end{figure}

\subsubsection{Improving Detection Performance}
Note that we have used Grounding DINO's pre-trained object concepts to detect objects in the drone image. However, we observed two issues with this approach that increased the number of false positives. First, the bounding boxes returned by the detection block often overlap with each other, resulting in over-prediction with the same piece of trash being detected multiple times. To reduce the over-prediction rate, Objects with significant overlap (40\% and higher) between the boxes that bounded them, were filtered and removed. 

The second issue involved classifying the entire image as an object. The drone image, captured at a 44.7-meter average altitude and at 5472 x 3648 pixel image resolution, could cover an area that contained many pieces of debris objects that would appear small in the image. Bounding boxes larger than half the size of the image were removed, to lower the mis-classification resulting from the entire image being detected as a debris object. These modification choices helped to customize the general-purpose zero-shot object detector to our specific problem domain without having to train our own model while minimizing issues related to redundancy and mis-classification. Figure~\ref{fig: misclassification} shows two examples of mis-classification in the object detection step.

\subsection{Classification using CLIP}

\subsubsection{Zero-shot classification} The second step in our pipeline is classifying the detected objects into relevant object categories. To classify the detected marine debris objects, we utilized CLIP~\cite{radford2021learning}, a zero-shot image classifier. CLIP uses a vision-language transformer pre-trained on image-text pairs, enabling general-purpose classification capability. As a result, given an image, the model can provide the most relevant text snippet describing the content of the image, without having to train a model. We selected CLIP due to its adaptability to classify new classes that are unseen during training using natural language instruction such as class names.  For this step, input images are created as image snippets cropped based on the bounding boxes in the object detection step. CLIP takes image and text pairs as input where text input is used as a list of possible class names for labeling the image. The class label with the highest probability is chosen as the final label for the object.

\subsubsection{Labeling schema for classification} To come up with a labeling schema for the classes that are appropriate for our specific problem domain, we relied on a manual marine debris audit conducted on the site where the drone images of this study were collected. The manual audit was performed by a group of researchers and volunteers as part of a community effort to track the spatial and temporal distribution of marine debris accumulation on coastal islands. We consulted their manual survey to develop a set of category labels that we can use as a natural language query for the CLIP model. Two of the authors further validated the choice of category labels by manually inspecting a random sub-sample of the images to merge, rename, and modify category names as necessary. We came up with 7 class categories: wood, cage, fishing gear, nature, plastic, metal, and wheel (see Table~\ref{tab:class_example} for examples from each class category).

\begin{table}[t!]
  \caption{Examples of debris objects for each class}
  \label{tab:trash}
  \begin{tabular}{ccl}
    \toprule
     Category Name & Example\\
    \midrule
    Wood &  Stray logs, boats\\
    Cage& Lobster traps \\
    Fishing Gear& Fishing nets, lobster buoys, ropes\\
   Nature & Bushes, rocks, moss, branches\\
    Plastic& Beverage bottles, caps, bags, cups\\
    Metal& Aluminium, tin cans\\
    Wheel& Tire, wheels\\
    \bottomrule
  \end{tabular}
  \Description{A table with example objects from the 7 debris object categories.}
  \label{tab:class_example}
\end{table}

\begin{figure}[t!] 
  \centering 
  \includegraphics[width=0.46\textwidth]{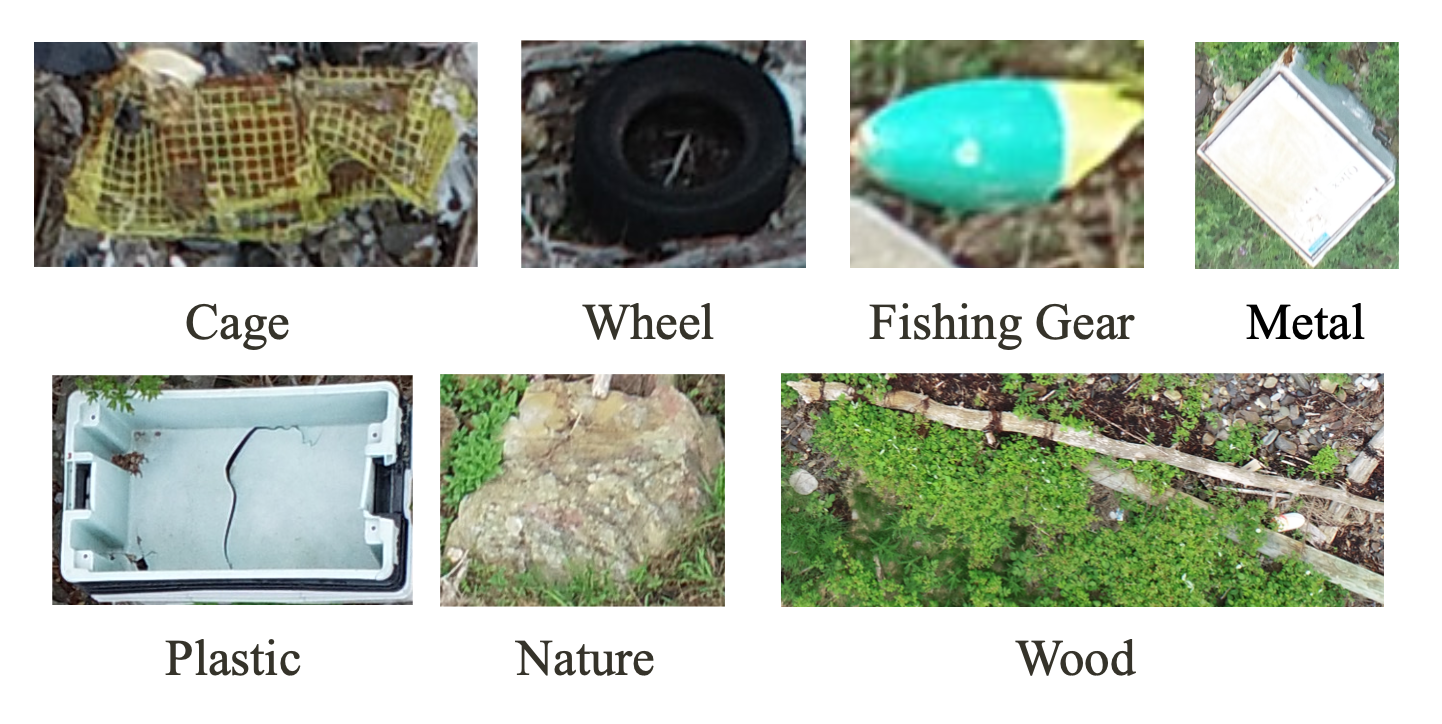}
  \caption{Example debris object from each of seven classes.}
  \Description{A figure with example objects from the 7 debris object categories.}
  \label{fig:example}
\end{figure}

Note that our labeling schema is inspired by commonly found marine debris on coastal islands. In particular, the economy of the areas where the data for this study was collected is dependent on fishing activities and a large quantity of the debris comes from fishing gear used by the fishing boats operating in the area. Another consideration when designing the labeling schema was to sort the detected materials for proper recycling, once collected. We consulted with representatives from the local community group volunteering in the marine debris removal process to understand how to design our framework to easily integrate it into their existing workflow. In the case of marine debris, some trash objects are heavier than others (each lobster trap weighs between 45 to 60 pounds), which often needs multiple volunteers to collect them. On the other hand, small plastic items (bottles, bags) can be retrieved by an individual volunteer easily. As a result, volunteers need to coordinate to ensure that they can allocate ample human resources to adequately cover the spatial distribution of the debris. Knowing the material type can help in this coordination as material type can be used as a good proxy for the expected weight of the object. Based on this finding, we designed the current labeling schema focusing on the material type to aid in the decision-making process of allocating resources such as volunteers and vehicles to both collect and transport the debris objects to recycling facilities located on the mainland. Figure~\ref{fig:example} shows example detection from each of the seven classes.

\subsection{Detecting duplicate objects}

Our drone images are captured using a predetermined flight path at a sampling rate of 30 frames per second. Due to the relatively high sampling rate, we observed that a particular piece of debris object appeared in multiple consecutive images resulting in duplicate items. To mitigate this issue, we developed a duplicate removal approach to avoid over-counting debris object instances. 

Our approach is based on two steps: 1) extracting distinctive features from the detected object, and 2) comparing different detected objects to identify identical objects and remove them. Note that during the flight, the drone can shift its orientation and thus the same piece of debris object can appear in different angles across consecutive overlapping images. To extract features from the detected objects, we utilized SIFT (Scale-Invariant Feature Transform), a computer vision algorithm that detects and describes local features in an image~\cite{lowe2004distinctive}. By identifying key points and descriptors in an object, SIFT extracts distinct features from each of the cropped images containing debris objects and checks whether they are different views of the same object. We chose SIFT for detecting the duplicates, as the SIFT key points are invariant to image scale, translation, rotation, and change in illumination, which is suitable to differentiate the different views of the same object in our problem setting.

One limitation of using SIFT is the algorithm's complexity, which can be prohibitive to use due to its slow speed and computational demand. To reduce the computational cost, we made several design choices. First, we decided to only compare an object to other candidate objects that are detected within a 5-meter radius distance of the query object. Second, to identify a match, SIFT requires setting a minimum number of key points. The higher the number of features considered, the more computationally expensive the process becomes. We set the minimum number of matching features to 50 based on our empirical experimentation. If the number of matching SIFT features exceeds 50, a pair of detected objects is considered to be a different view of the same object. As SIFT has a runtime complexity of $O(MN + I)$ for $M*N$ image and $I$ number of descriptors, these two practical strategies reduced the computational complexity of our proposed system.

\subsection{Map visualization}

\subsubsection{Goal} Our objective is to aid in the coordination of volunteer-led marine debris collection efforts through remote detection and mapping of debris using our system. Based on our consultation experience with local stakeholders, volunteer teams are on the island for a scheduled day (usually weekends) and spend the daytime hours trying to cover as many sites as possible. 
Consecutive years' manual auditing reports can help identify the sites with high-density debris objects. Otherwise, teams of volunteers have to manually survey the island on foot, which can often take hours depending on the terrain conditions, weather, and other factors. When volunteers reach the island, knowing information such as where the hotspots are with marine debris located in high density, whether the objects are light or heavy, etc. can help the volunteer group to assign teams of people to those specific locations to utilize their time and effort more effectively. For this purpose, we focused on creating a map visualization of the drone survey that can be made available to the volunteer team to streamline the removal process.

We used the metadata information from the drone image (longitude, latitude, and altitude) to obtain the real-world point coordinates of the detected objects. 
To map the detected objects to their corresponding location, the spatial resolution can be estimated with the dimension of the bounding box for a detected object.
To do this, we needed to calculate the Ground Sample Distance (GSD) for each image location based on the corresponding altitude and camera intrinsic parameters to capture the details. 

\subsubsection{Ground Sample Distance} In remote sensing from satellite or drone images, each square pixel represents a square land or sea area on the earth's surface. Ground Sampling Distance (GSD) is a measure used in remote sensing to define the linear distance of a single pixel in an image. It describes how much ground area is represented by one pixel and is typically expressed in meters. The larger the GSD is, the lower spatial resolution is gained. GSD allows us to estimate the real-world distance from the image-view distance. 

\begin{figure*}[t]
  \centering
  \includegraphics[width=0.90\textwidth]{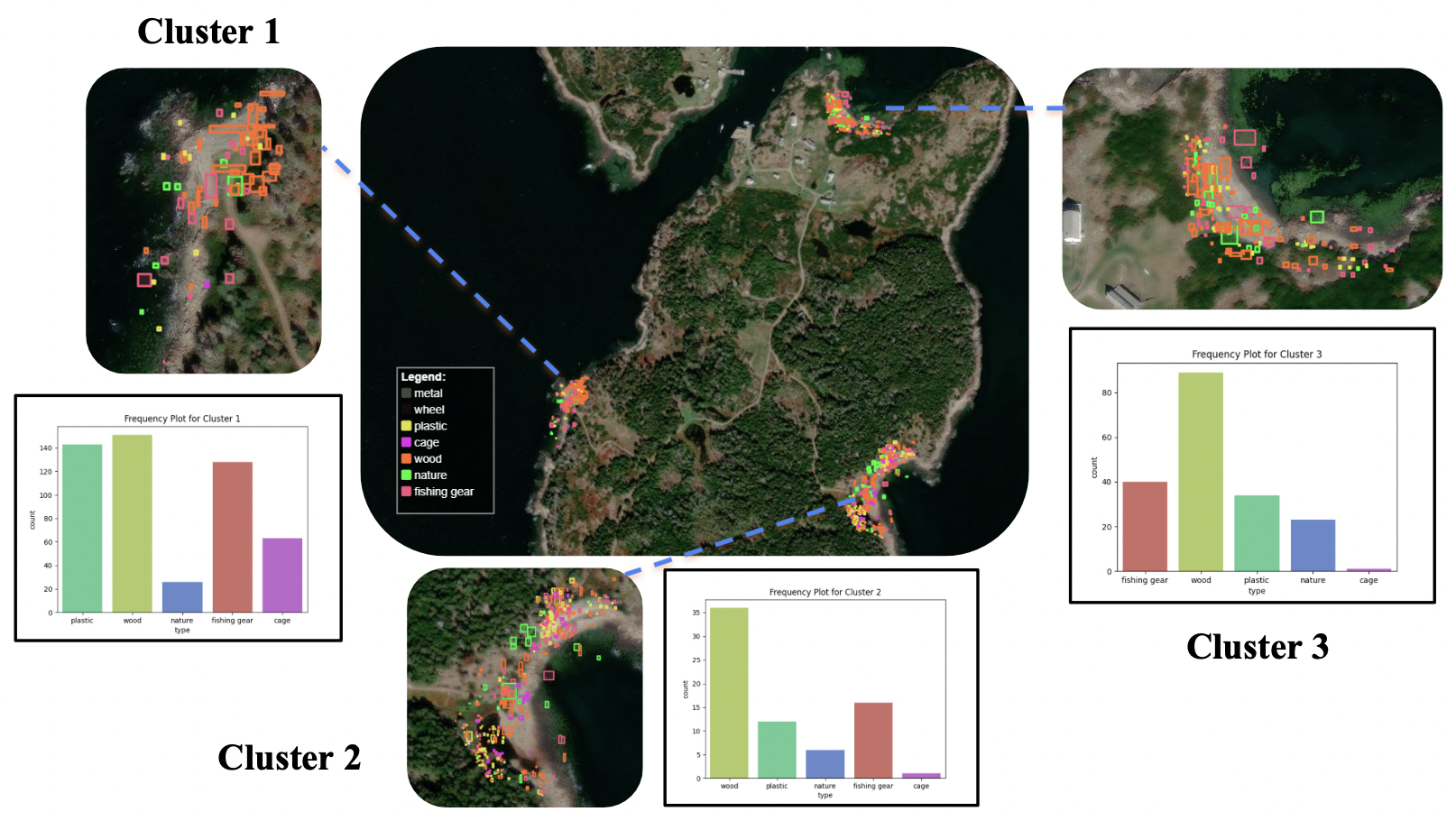}
  \caption{Detected clusters of marine debris distribution.}
\Description{Detected clusters of marine debris distribution on the island.}
\label{fig:interface}
\end{figure*}

We leveraged the following relationship for GSD estimation,
\begin{equation}\label{equation 1}
\text{Ground Sample Distance} = \frac{{A \cdot S}}{{f \cdot I}}
\end{equation}

where, \begin{itemize}
\item \( A \): Altitude of the sensor (Height between the camera and the ground) in meters.
\item \( S \): Sensor width (in meters).
\item \( f \): Focal length of the camera lens (in millimeters).
\item \( I \): Image width (in meters).
\end{itemize}

As the focal length of our drone is fixed, we use altitude, the object's image-view, and real-world dimension to estimate the GSD from equation~\ref{equation 1}. Using GSD, we then translate the image-level center position of each object to its real-world position and map them to corresponding geo-spatial coordinates. An example of the map is presented in Figure~\ref{fig:interface}, where three clusters of debris objects can be seen.

One limitation we encountered in GSD calculation is camera angle and variable GSD due to altitude change during drone flight. Due to the tree height and ground elevation change, drone altitude during the flight dynamically changed for safe navigation. On the other hand, lower altitude results in fine-grained resolution, but result in more time and images required to survey a specific area. These practical considerations are important design choices for deploying the system in longitudinal automated surveys and its performance, which we plan to explore in our future work. 


\section{Evaluation}

\subsection{Site}

To validate the performance of the proposed framework, we use a new drone dataset of marine debris images. The dataset was primarily collected on Allen Island, a coastal island located in North America of 2 km$^2$ area. The island is owned by Colby College and is primarily used for research and educational purposes owing to the rich vegetation and multiple beach substrates, that are ideal for studying different beach types. Visitors and volunteers can reach the island primarily by boat, especially during summer. High fishing and tourist activities in the water close to the island result in a large number of marine debris objects accumulating along the shore in every fishing season, making it ideal for the experimental evaluation of our framework.


\subsection{Data Collection}

Our dataset comprised of 1,000 images. The images were sampled from a single drone flight surveying the island landscape on May 30, 2022. The aerial survey was performed by a DJI Phantom 4 Pro (maximum image dimension 5472 by 3648).  

While our framework is designed to perform without human labels as we do not train or modify an existing detection model, we needed to curate a labeled dataset with human-provided marine debris category labels for evaluation purposes. The dataset was manually labeled by 6 undergraduate students as part of a class project utilizing LabelBox~\cite{LabeBox}, a web-based labeling tool freely available for research purposes. 200 images were randomly selected as an initial set that the annotators labeled individually. To minimize human bias in the labeling process, an annotation guide~\cite{tseng2020best} was developed through iterative discussion among the annotators to ensure consistency in the labels across the annotators. Following the annotation guide, the human annotators labeled the entire dataset using a labeling schema with the final class categories.

\begin{table}[t!] 
  \caption{Detection and Classification Performance Evaluation}
  \label{tab:performnace}
  \begin{tabular}{ccl}
    \toprule
    Task & Metric & Score\\
    \midrule
    \texttt Object Detection & Mean IOU & 0.69 \\
    \texttt Object Classification & F1  & 0.74 \\
    \bottomrule
  \end{tabular}
\end{table}

\subsection{Results}

We evaluate the detection and classification performance of our framework using the annotator's labels as ground truth. Since our framework involves two modules: object detection and classification that are interdependent, we evaluate the performance of each module separately. Note that, while interdependent, performance in each module can be of value for clean-up efforts, as a correctly detected debris objects that is categorized to a wrong class can still help the volunteers. 

For object detection, we use mean Intersection over Union (IOU), which quantifies the overlap between the predicted bounding box and the ground truth bounding box. The score ranges between 0 to 1 where a higher score indicates better performance. For the classification step, we use the F1 score, which is the harmonic mean of precision (fraction of correctly detected objects from total detection) and recall (fraction of correctly detected objects from total objects), ranging between 0  to 1 (best performance). The F1 score allows one to measure the model's performance by taking misclassifications into account, which is particularly useful for evaluating the overall effectiveness of the model when the number of instances from different classes is imbalanced.  

Note that while prior works exist where marine debris objects are detected and classified using computer vision models~\cite{WINANS2023103515, kikaki2022marida, marin2021deep}, comparing our results directly with those systems can result in spurious conclusions. First, most prior works trained or finetuned a model using human labels in a supervised learning setting. Second, different marine debris detection methods involves different numbers and types of object categories depending on the economic activities of the data collection site. 

\subsubsection{Detection performance} The evaluation results are presented in Table~\ref{tab:performnace}. The mean IoU score for marine debris detection using Grounding DINO is 0.69. 
This is comparable to the performance reported with existing drone-based deep learning-based litter detection approaches~\cite{rs13050965,  politikos2021automatic}. This is promising considering we are using no human labels in detection and the objects of interest are different from general waste categories. The detection performance can be further improved as our database of correctly detected marine debris objects grows to be large enough to train or finetune an existing object detector using those labels as ground truth.

\subsubsection{Classification performance} Our CLIP based marine debris classifier reported an accuracy of $0.775$ and average F1 score of $ 0.74$6 across the 7 identified classes. Our F1 score is close to previous works on marine debris detection~\cite{WINANS2023103515, drones6120401} using human labels, reporting similar (around 0.72 F1 score) performance. This is promising as we are operating on multi-class classification without any human labels.

One approach to improve the classification performance is dependent on detection performance. Our objects of interest are of varying sizes and are often found in clusters resulting in overlapping bounding boxes. Smaller items such as bottles and caps were challenging to identify and label separately for human annotators. This lowers the amount of overlap between ground truth and detection, affecting the classification performance. ~\cite{WINANS2023103515} found results suggesting that the performance of marine debris detection from drone images improves with pixel-per-object ratio. We believe that systematic exploration of image resolution and target debris size can be useful to further improve performance.

\section{Web interface}

Our objective is to provide the user easy access to the spatial distribution information about the debris object so that it can aid in the removal process. Towards that end, we developed a web application to allow stakeholders to easily use the framework and explore the results. The front end of the application is developed using React javascript library~\cite{React} that serves as a Flask application. Our web interface is accessible as a public Docker container, which is available at: \url{https://github.com/Tahiya31/Trash_Track}.

\subsection{User Interaction}
The web interface is designed to offer an intuitive user experience targeting non-technical stakeholders. Based on the findings of our need-finding session with volunteers involved in removal and coordination, the application workflow involves six functionality pages: Home, Data, Duplicate, Map, Plot, and Help.

\begin{figure}[t]
  \centering
  \includegraphics[width=0.95\linewidth]{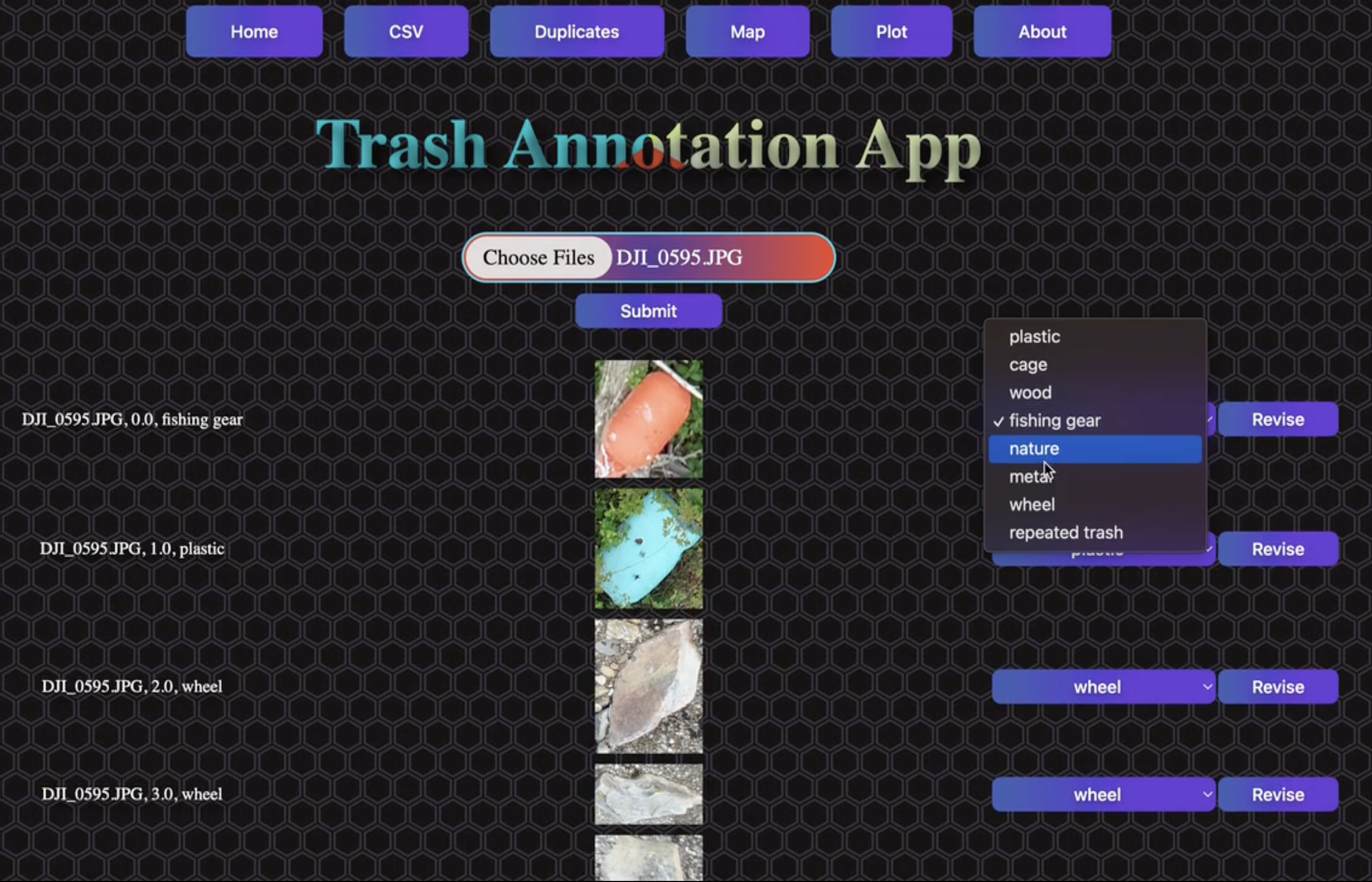}
  \caption{A screen shot of the Home page of the web app. User can build an annotated dataset for their custom geographic region by collaboratively annotating using computer vision models available through our framework.}
\label{fig:web_app}
\end{figure}

\textit{Home page.} On this page, users can upload a set of drone images. Once uploaded, the user can run object detection and classification modules on each image. Every detected debris object is cropped and displayed with the predicted label. If incorrect, the user can manually correct the class label from a drop-down menu containing the 7 predefined classes, allowing the user to iteratively improve the model-provided annotation and create a labeled dataset. Figure~\ref{fig:web_app} shows the Home page where a drop-down menu is used to correct the predicted label of a debris item.

\textit{Data.} The Data page allows the user to upload, view, remove,  and download the current database containing the results in a tabular format. All user interaction including label corrections made within the web application is stored as a .csv file that the user can download for their analysis. The .csv file contains bounding box coordinates, image metadata (latitude, longitude, altitude), and class label information, which can be used for additional analysis.

\textit{Duplicates.} Users can upload a set of images here, and check whether there are duplicate objects. If found, duplicate objects are removed from database.

\textit{Map.} This page presents all detected debris items on a map, with detected trash overlayed on it corresponding to their real-world position. Debris object types are coded by specific colors to aid in the removal.

\textit{Plot.} The Plot page generates charts and plots to quickly visualize the distribution of different trash categories to understand the dominant category on a particular site and aid in volunteer management and removal planning.

We also included a \textit{Help} page with instructions and a demonstration with a video walk-through illustrating the application's different functionalities described above.


\section{Ethical Considerations and Limitations}

The framework proposed in this work is designed using multiple machine learning-based models to detect and classify objects. These models are trained on large datasets for generalized performance that are primarily well-suited for western geographic areas and thus may not be well-performant in other geographic regions. Relying solely on automated tools for detecting objects in critical settings can thus lead to potential societal and environmental consequences due to misclassification.

Our drone images are collected on a remote island with very few inhabitants. Drone images, taken in residential areas, can contain individually identifiable information and can invade people's privacy. Another limitation is our framework is designed and evaluated on data collected from a single geographic region. While we have tried to ensure that our framework is built on general-purpose detection and classification models to be useful in a wide variety of custom marine debris collection applications, we have not empirically evaluated the approach for data collected in other sites, which we plan to do in the future.

We utilized zero-shot object detectors and classifiers, which are inherently large models trained on millions of training images and are computationally expensive during inference time. Another source of computational expense in this framework is the duplicate matching step with SIFT, which has a quadratic time complexity in a worst-case scenario. Our current framework is supported by Paperspace Gradient, a high-performance computing platform, useful for prototyping with an hourly pay-as-you-go pricing model. In the future, we plan to find more economic alternatives for hosting the web application and running the model for real-world deployment.

\section{Discussion}

While detecting and classifying waste objects with aerial imagery have been explored in existing works, most approaches rely on publicly available large-scale datasets. However, real-world marine debris cleanup efforts need user-friendly frameworks that can be easily adaptable with minimal annotation efforts by communities from different geographic landscapes and environmental conditions. As our focus is to develop a solution that can be sustainably adopted to existing clean-up efforts, we describe these considerations through additional data collection and future directions to improve its impact.

\subsection{Data Collection Considerations}

We ran a pilot data collection effort to gather a second drone dataset of marine debris images in January 2024. We picked several beaches along islands in Bermuda for this pilot study. Bermuda beaches have a similar marine waste pollution problem where beaches are covered with millions of small pieces of plastic trash~\cite{bermuda}. These trash pieces come from the Sargasso Sea, which is bounded by multiple currents in the Atlantic that carry the trash into the ocean and eventually accumulate on islands like the ones in Bermuda due to the weather pattern.

We experienced that environmental factors can impact such drone data collection tremendously. The attempt to gather data in Bermuda provided valuable insights and underscored several challenges inherent to deploying this type of technology in new geographic areas and environments.

First, the quality of images captured in a drone flight is subject to weather conditions. Adverse weather conditions, including high winds and sudden rain, disrupted the flight path of the drone and its ability to maintain a stable altitude position that is necessary to capture high-resolution images. Second, collecting drone images on public land and beaches was often restricted as they were categorized as no-fly zones by local government and regulatory bodies. Due to these complications and budgetary restrictions, it was not possible to collect adequate data in Bermuda for evaluating our framework. Based on the lessons learned from this experience, we plan to organize a data collection plan in the future by working with local stakeholders to identify suitable sites and weather conditions to appropriately cover different marine debris hotspots with various types and number of object classes.

\subsection{Future Works}

We have presented preliminary works toward designing a framework for sustainable marine debris removal through community-driven efforts. Currently, our framework performs without human labels using a zero-shot object detector (Grounding DINO) and object classifier (CLIP). While ML models are assessed by their accuracy on the task against ground truth labels, the energy efficiency and computational expense are also important factors for budgetary decisions of community-led efforts. In the future, more experiments are needed to assess the performance and efficiency of the framework using other detection and classification models such as OWL~\cite{minderer2024scaling}, that can help to scale this framework to a wide range of marine debris objects in open-world classification setting. We also want to evaluate the applicability of the framework for different types of marine debris contexts and coastal ecosystems to validate its robustness.


We initiated this project to explore building a community-driven solution for the marine debris problem in small remote islands. The current approach of frequently surveying the islands manually requires significant budget allocation and volunteer time, which does not scale well. In the future, we plan to release the framework as a mobile application that visitors and volunteers of the island can download to access the trash locations for ease of planning. We also plan to add functionality to the app where application users can report the location of a debris item if it has not been located by the system (e.g. occluded from drone view) already. This would also act as the ground truth for assessing the detection and classification model's performance in future iterations. Debris items collected in this manner will be sorted based on their recycling potential and can be transported to a material recovery facility to maximize the value of recycled materials. 


Realizing the positive impact of technological interventions often depends on local stakeholders' knowledge. For example, fishing gear that get washed up are still considered the fishermen's private property in many coastal regions, making it required for volunteers to return the collected gears to their owners after removal. Volunteers and fishermen work together to organize a `Gear Grab' event to ensure gears are returned to their rightful owners and remove other debris items for recycling. Additionally, storms and hurricanes play a role in displacing the objects from their located position, often moving them inward into the woods with the wind and waves. To bridge the gap between technological innovation and on-ground efforts, we aim to 
collaborate with volunteers and organizations locally involved in on-ground clean up efforts. Such collaboration can provide practical insights and feedback on such issues that can improve the user experience, usability, efficiency, and overall performance of our framework in supporting clean-up initiatives. This engagement will facilitate iterative improvements to the system, ensuring it meets the practical needs of users and maximizes the sustainable conservation of island ecosystems.

\section{Conclusion}

In this work, we have introduced a framework for marine debris cleanup initiatives led by the local government and volunteer community. Our framework is designed to reduce the human efforts involved in the current manual survey method and expedite the coordination of the debris collection and removal process. This framework uses state-of-the-art object detection and classification approaches to automate the monitoring process that stakeholders can easily use and visualize through an interactive web application. We also evaluate our approach using human-annotated drone images from a novel dataset of marine debris on a coastal island. The proposed approach can be used without human annotations for custom model training and can thus be used for a wide range of geographic regions and environmental conditions. Our framework is publicly available that we hope will enable sustainable community-driven efforts for cleaning up our oceans and coast.


\begin{acks}
We thank Dr. Amanda Stent from the Davis Institute for Artificial Intelligence for her input in the annotation and initial framework design. We also thank 
Tim Stonesifer for drone data collection, manual survey, and feedback on the framework design. We acknowledge the students in Colby January Course JP297 for helping with the annotation of the dataset.
\end{acks}

\bibliographystyle{ACM-Reference-Format}
\bibliography{ref}


\begin{thebibliography}{25}


\ifx \showCODEN    \undefined \def \showCODEN     #1{\unskip}     \fi
\ifx \showDOI      \undefined \def \showDOI       #1{#1}\fi
\ifx \showISBNx    \undefined \def \showISBNx     #1{\unskip}     \fi
\ifx \showISBNxiii \undefined \def \showISBNxiii  #1{\unskip}     \fi
\ifx \showISSN     \undefined \def \showISSN      #1{\unskip}     \fi
\ifx \showLCCN     \undefined \def \showLCCN      #1{\unskip}     \fi
\ifx \shownote     \undefined \def \shownote      #1{#1}          \fi
\ifx \showarticletitle \undefined \def \showarticletitle #1{#1}   \fi
\ifx \showURL      \undefined \def \showURL       {\relax}        \fi
\providecommand\bibfield[2]{#2}
\providecommand\bibinfo[2]{#2}
\providecommand\natexlab[1]{#1}
\providecommand\showeprint[2][]{arXiv:#2}

\bibitem[UNE(2022)]%
        {UNEPreport}
 \bibinfo{year}{2022}\natexlab{}.
\newblock \bibinfo{booktitle}{\emph{Our Planet is choking on plastic}}.
\newblock
\urldef\tempurl%
\url{https://www.unep.org/interactives/beat-plastic-pollution/}
\showURL{%
Retrieved March 7, 2024 from \tempurl}


\bibitem[Lab(2023)]%
        {LabeBox}
 \bibinfo{year}{2023}\natexlab{}.
\newblock \bibinfo{booktitle}{\emph{LabelBox}}.
\newblock
\urldef\tempurl%
\url{https://labelbox.com}
\showURL{%
Retrieved March 5, 2024 from \tempurl}


\bibitem[Rea(2023)]%
        {React}
 \bibinfo{year}{2023}\natexlab{}.
\newblock \bibinfo{booktitle}{\emph{React}}.
\newblock
\urldef\tempurl%
\url{https://react.dev}
\showURL{%
Retrieved March 5, 2024 from \tempurl}


\bibitem[Finighan(2024)]%
        {bermuda}
\bibfield{author}{\bibinfo{person}{Gareth Finighan}.}
  \bibinfo{year}{2024}\natexlab{}.
\newblock \bibinfo{booktitle}{\emph{Ocean pollution crisis is ‘literally
  choking our local marine environments’ – KBB}}.
\newblock
\urldef\tempurl%
\url{https://www.royalgazette.com/environment/news/article/20230109/ocean-pollution-crisis-is-literally-choking-our-local-marine-environments-kbb/}
\showURL{%
Retrieved March 9, 2024 from \tempurl}


\bibitem[Freitas et~al\mbox{.}(2021)]%
        {freitas2021remote}
\bibfield{author}{\bibinfo{person}{Sara Freitas}, \bibinfo{person}{Hugo Silva},
  {and} \bibinfo{person}{Eduardo Silva}.} \bibinfo{year}{2021}\natexlab{}.
\newblock \showarticletitle{Remote hyperspectral imaging acquisition and
  characterization for marine litter detection}.
\newblock \bibinfo{journal}{\emph{Remote Sensing}} \bibinfo{volume}{13},
  \bibinfo{number}{13} (\bibinfo{year}{2021}), \bibinfo{pages}{2536}.
\newblock


\bibitem[Garaba et~al\mbox{.}(2020)]%
        {essd-12-2665-2020}
\bibfield{author}{\bibinfo{person}{S.~P. Garaba}, \bibinfo{person}{T. Acu\~na
  Ruz}, {and} \bibinfo{person}{C.~B. Mattar}.} \bibinfo{year}{2020}\natexlab{}.
\newblock \showarticletitle{Hyperspectral longwave infrared reflectance spectra
  of naturally dried algae, anthropogenic plastics, sands and shells}.
\newblock \bibinfo{journal}{\emph{Earth System Science Data}}
  \bibinfo{volume}{12}, \bibinfo{number}{4} (\bibinfo{year}{2020}),
  \bibinfo{pages}{2665--2678}.
\newblock
\urldef\tempurl%
\url{https://doi.org/10.5194/essd-12-2665-2020}
\showDOI{\tempurl}


\bibitem[Iordache et~al\mbox{.}(2022)]%
        {iordache2022targeting}
\bibfield{author}{\bibinfo{person}{Marian-Daniel Iordache},
  \bibinfo{person}{Liesbeth De~Keukelaere}, \bibinfo{person}{Robrecht Moelans},
  \bibinfo{person}{Lisa Landuyt}, \bibinfo{person}{Mehrdad Moshtaghi},
  \bibinfo{person}{Paolo Corradi}, {and} \bibinfo{person}{Els Knaeps}.}
  \bibinfo{year}{2022}\natexlab{}.
\newblock \showarticletitle{Targeting plastics: machine learning applied to
  litter detection in aerial multispectral images}.
\newblock \bibinfo{journal}{\emph{Remote Sensing}} \bibinfo{volume}{14},
  \bibinfo{number}{22} (\bibinfo{year}{2022}), \bibinfo{pages}{5820}.
\newblock


\bibitem[Kikaki et~al\mbox{.}(2022)]%
        {kikaki2022marida}
\bibfield{author}{\bibinfo{person}{Katerina Kikaki}, \bibinfo{person}{Ioannis
  Kakogeorgiou}, \bibinfo{person}{Paraskevi Mikeli},
  \bibinfo{person}{Dionysios~E Raitsos}, {and} \bibinfo{person}{Konstantinos
  Karantzalos}.} \bibinfo{year}{2022}\natexlab{}.
\newblock \showarticletitle{MARIDA: A benchmark for Marine Debris detection
  from Sentinel-2 remote sensing data}.
\newblock \bibinfo{journal}{\emph{PloS one}} \bibinfo{volume}{17},
  \bibinfo{number}{1} (\bibinfo{year}{2022}), \bibinfo{pages}{e0262247}.
\newblock


\bibitem[Knaeps et~al\mbox{.}(2021)]%
        {essd-13-713-2021}
\bibfield{author}{\bibinfo{person}{E. Knaeps}, \bibinfo{person}{S. Sterckx},
  \bibinfo{person}{G. Strackx}, \bibinfo{person}{J. Mijnendonckx},
  \bibinfo{person}{M. Moshtaghi}, \bibinfo{person}{S.~P. Garaba}, {and}
  \bibinfo{person}{D. Meire}.} \bibinfo{year}{2021}\natexlab{}.
\newblock \showarticletitle{Hyperspectral-reflectance dataset of dry, wet and
  submerged marine litter}.
\newblock \bibinfo{journal}{\emph{Earth System Science Data}}
  \bibinfo{volume}{13}, \bibinfo{number}{2} (\bibinfo{year}{2021}),
  \bibinfo{pages}{713--730}.
\newblock
\urldef\tempurl%
\url{https://doi.org/10.5194/essd-13-713-2021}
\showDOI{\tempurl}


\bibitem[Kraft et~al\mbox{.}(2021)]%
        {rs13050965}
\bibfield{author}{\bibinfo{person}{Marek Kraft}, \bibinfo{person}{Mateusz
  Piechocki}, \bibinfo{person}{Bartosz Ptak}, {and} \bibinfo{person}{Krzysztof
  Walas}.} \bibinfo{year}{2021}\natexlab{}.
\newblock \showarticletitle{Autonomous, Onboard Vision-Based Trash and Litter
  Detection in Low Altitude Aerial Images Collected by an Unmanned Aerial
  Vehicle}.
\newblock \bibinfo{journal}{\emph{Remote Sensing}} \bibinfo{volume}{13},
  \bibinfo{number}{5} (\bibinfo{year}{2021}).
\newblock
\showISSN{2072-4292}
\urldef\tempurl%
\url{https://doi.org/10.3390/rs13050965}
\showDOI{\tempurl}


\bibitem[Lincoln et~al\mbox{.}(2022)]%
        {LINCOLN2022155709}
\bibfield{author}{\bibinfo{person}{Susana Lincoln}, \bibinfo{person}{Barnaby
  Andrews}, \bibinfo{person}{Silvana~N.R. Birchenough}, \bibinfo{person}{Piyali
  Chowdhury}, \bibinfo{person}{Georg~H. Engelhard}, \bibinfo{person}{Olivia
  Harrod}, \bibinfo{person}{John~K. Pinnegar}, {and} \bibinfo{person}{Bryony~L.
  Townhill}.} \bibinfo{year}{2022}\natexlab{}.
\newblock \showarticletitle{Marine litter and climate change: Inextricably
  connected threats to the world's oceans}.
\newblock \bibinfo{journal}{\emph{Science of The Total Environment}}
  \bibinfo{volume}{837} (\bibinfo{year}{2022}), \bibinfo{pages}{155709}.
\newblock
\showISSN{0048-9697}
\urldef\tempurl%
\url{https://doi.org/10.1016/j.scitotenv.2022.155709}
\showDOI{\tempurl}


\bibitem[Liu et~al\mbox{.}(2023)]%
        {liu2023grounding}
\bibfield{author}{\bibinfo{person}{Shilong Liu}, \bibinfo{person}{Zhaoyang
  Zeng}, \bibinfo{person}{Tianhe Ren}, \bibinfo{person}{Feng Li},
  \bibinfo{person}{Hao Zhang}, \bibinfo{person}{Jie Yang},
  \bibinfo{person}{Chunyuan Li}, \bibinfo{person}{Jianwei Yang},
  \bibinfo{person}{Hang Su}, \bibinfo{person}{Jun Zhu}, {et~al\mbox{.}}}
  \bibinfo{year}{2023}\natexlab{}.
\newblock \showarticletitle{Grounding dino: Marrying dino with grounded
  pre-training for open-set object detection}.
\newblock \bibinfo{journal}{\emph{arXiv preprint arXiv:2303.05499}}
  (\bibinfo{year}{2023}).
\newblock


\bibitem[Lowe(2004)]%
        {lowe2004distinctive}
\bibfield{author}{\bibinfo{person}{David~G Lowe}.}
  \bibinfo{year}{2004}\natexlab{}.
\newblock \showarticletitle{Distinctive image features from scale-invariant
  keypoints}.
\newblock \bibinfo{journal}{\emph{International journal of computer vision}}
  \bibinfo{volume}{60} (\bibinfo{year}{2004}), \bibinfo{pages}{91--110}.
\newblock


\bibitem[Marin et~al\mbox{.}(2021)]%
        {marin2021deep}
\bibfield{author}{\bibinfo{person}{Ivana Marin}, \bibinfo{person}{Sa{\v{s}}a
  Mladenovi{\'c}}, \bibinfo{person}{Sven Gotovac}, {and} \bibinfo{person}{Goran
  Zaharija}.} \bibinfo{year}{2021}\natexlab{}.
\newblock \showarticletitle{Deep-feature-based approach to marine debris
  classification}.
\newblock \bibinfo{journal}{\emph{Applied Sciences}} \bibinfo{volume}{11},
  \bibinfo{number}{12} (\bibinfo{year}{2021}), \bibinfo{pages}{5644}.
\newblock


\bibitem[Martin et~al\mbox{.}(2021)]%
        {martin2021enabling}
\bibfield{author}{\bibinfo{person}{Cecilia Martin}, \bibinfo{person}{Qiannan
  Zhang}, \bibinfo{person}{Dongjun Zhai}, \bibinfo{person}{Xiangliang Zhang},
  {and} \bibinfo{person}{Carlos~M Duarte}.} \bibinfo{year}{2021}\natexlab{}.
\newblock \showarticletitle{Enabling a large-scale assessment of litter along
  Saudi Arabian red sea shores by combining drones and machine learning}.
\newblock \bibinfo{journal}{\emph{Environmental Pollution}}
  \bibinfo{volume}{277} (\bibinfo{year}{2021}), \bibinfo{pages}{116730}.
\newblock


\bibitem[Merlino et~al\mbox{.}(2021)]%
        {merlino2021citizen}
\bibfield{author}{\bibinfo{person}{Silvia Merlino}, \bibinfo{person}{Marco
  Paterni}, \bibinfo{person}{Marina Locritani}, \bibinfo{person}{Umberto
  Andriolo}, \bibinfo{person}{Gil Gon{\c{c}}alves}, {and}
  \bibinfo{person}{Luciano Massetti}.} \bibinfo{year}{2021}\natexlab{}.
\newblock \showarticletitle{Citizen science for marine litter detection and
  classification on unmanned aerial vehicle images}.
\newblock \bibinfo{journal}{\emph{Water}} \bibinfo{volume}{13},
  \bibinfo{number}{23} (\bibinfo{year}{2021}), \bibinfo{pages}{3349}.
\newblock


\bibitem[Minderer et~al\mbox{.}(2024)]%
        {minderer2024scaling}
\bibfield{author}{\bibinfo{person}{Matthias Minderer}, \bibinfo{person}{Alexey
  Gritsenko}, {and} \bibinfo{person}{Neil Houlsby}.}
  \bibinfo{year}{2024}\natexlab{}.
\newblock \showarticletitle{Scaling open-vocabulary object detection}.
\newblock \bibinfo{journal}{\emph{Advances in Neural Information Processing
  Systems}}  \bibinfo{volume}{36} (\bibinfo{year}{2024}).
\newblock


\bibitem[Moy et~al\mbox{.}(2018)]%
        {moy2018mapping}
\bibfield{author}{\bibinfo{person}{Kirsten Moy}, \bibinfo{person}{Brian
  Neilson}, \bibinfo{person}{Anne Chung}, \bibinfo{person}{Amber Meadows},
  \bibinfo{person}{Miguel Castrence}, \bibinfo{person}{Stephen Ambagis}, {and}
  \bibinfo{person}{Kristine Davidson}.} \bibinfo{year}{2018}\natexlab{}.
\newblock \showarticletitle{Mapping coastal marine debris using aerial imagery
  and spatial analysis}.
\newblock \bibinfo{journal}{\emph{Marine pollution bulletin}}
  \bibinfo{volume}{132} (\bibinfo{year}{2018}), \bibinfo{pages}{52--59}.
\newblock


\bibitem[Pfeiffer et~al\mbox{.}(2022)]%
        {pfeiffer2022use}
\bibfield{author}{\bibinfo{person}{Roland Pfeiffer}, \bibinfo{person}{Gianluca
  Valentino}, \bibinfo{person}{Sebastiano D’Amico}, \bibinfo{person}{Luca
  Piroddi}, \bibinfo{person}{Luciano Galone}, \bibinfo{person}{Stefano
  Calleja}, \bibinfo{person}{Reuben~A Farrugia}, {and}
  \bibinfo{person}{Emanuele Colica}.} \bibinfo{year}{2022}\natexlab{}.
\newblock \showarticletitle{Use of UAVs and Deep Learning for Beach Litter
  Monitoring}.
\newblock \bibinfo{journal}{\emph{Electronics}} \bibinfo{volume}{12},
  \bibinfo{number}{1} (\bibinfo{year}{2022}), \bibinfo{pages}{198}.
\newblock


\bibitem[Politikos et~al\mbox{.}(2021)]%
        {politikos2021automatic}
\bibfield{author}{\bibinfo{person}{Dimitris~V Politikos},
  \bibinfo{person}{Elias Fakiris}, \bibinfo{person}{Athanasios Davvetas},
  \bibinfo{person}{Iraklis~A Klampanos}, {and} \bibinfo{person}{George
  Papatheodorou}.} \bibinfo{year}{2021}\natexlab{}.
\newblock \showarticletitle{Automatic detection of seafloor marine litter using
  towed camera images and deep learning}.
\newblock \bibinfo{journal}{\emph{Marine Pollution Bulletin}}
  \bibinfo{volume}{164} (\bibinfo{year}{2021}), \bibinfo{pages}{111974}.
\newblock


\bibitem[Radford et~al\mbox{.}(2021)]%
        {radford2021learning}
\bibfield{author}{\bibinfo{person}{Alec Radford}, \bibinfo{person}{Jong~Wook
  Kim}, \bibinfo{person}{Chris Hallacy}, \bibinfo{person}{Aditya Ramesh},
  \bibinfo{person}{Gabriel Goh}, \bibinfo{person}{Sandhini Agarwal},
  \bibinfo{person}{Girish Sastry}, \bibinfo{person}{Amanda Askell},
  \bibinfo{person}{Pamela Mishkin}, \bibinfo{person}{Jack Clark},
  {et~al\mbox{.}}} \bibinfo{year}{2021}\natexlab{}.
\newblock \showarticletitle{Learning transferable visual models from natural
  language supervision}. In \bibinfo{booktitle}{\emph{International conference
  on machine learning}}. PMLR, \bibinfo{pages}{8748--8763}.
\newblock


\bibitem[Tasseron et~al\mbox{.}(2021)]%
        {tasseron2021advancing}
\bibfield{author}{\bibinfo{person}{Paolo Tasseron}, \bibinfo{person}{Tim
  Van~Emmerik}, \bibinfo{person}{Joseph Peller}, \bibinfo{person}{Louise
  Schreyers}, {and} \bibinfo{person}{Lauren Biermann}.}
  \bibinfo{year}{2021}\natexlab{}.
\newblock \showarticletitle{Advancing floating macroplastic detection from
  space using experimental hyperspectral imagery}.
\newblock \bibinfo{journal}{\emph{Remote Sensing}} \bibinfo{volume}{13},
  \bibinfo{number}{12} (\bibinfo{year}{2021}), \bibinfo{pages}{2335}.
\newblock


\bibitem[Tran et~al\mbox{.}(2022)]%
        {drones6120401}
\bibfield{author}{\bibinfo{person}{Thi Linh~Chi Tran},
  \bibinfo{person}{Zhi-Cheng Huang}, \bibinfo{person}{Kuo-Hsin Tseng}, {and}
  \bibinfo{person}{Ping-Hsien Chou}.} \bibinfo{year}{2022}\natexlab{}.
\newblock \showarticletitle{Detection of Bottle Marine Debris Using Unmanned
  Aerial Vehicles and Machine Learning Techniques}.
\newblock \bibinfo{journal}{\emph{Drones}} \bibinfo{volume}{6},
  \bibinfo{number}{12} (\bibinfo{year}{2022}).
\newblock
\showISSN{2504-446X}
\urldef\tempurl%
\url{https://doi.org/10.3390/drones6120401}
\showDOI{\tempurl}


\bibitem[Tseng et~al\mbox{.}(2020)]%
        {tseng2020best}
\bibfield{author}{\bibinfo{person}{Tina Tseng}, \bibinfo{person}{Amanda Stent},
  {and} \bibinfo{person}{Domenic Maida}.} \bibinfo{year}{2020}\natexlab{}.
\newblock \showarticletitle{Best practices for managing data annotation
  projects}.
\newblock \bibinfo{journal}{\emph{arXiv preprint arXiv:2009.11654}}
  (\bibinfo{year}{2020}).
\newblock


\bibitem[Winans et~al\mbox{.}(2023)]%
        {WINANS2023103515}
\bibfield{author}{\bibinfo{person}{W.~Ross Winans}, \bibinfo{person}{Qi Chen},
  \bibinfo{person}{Yi Qiang}, {and} \bibinfo{person}{Erik~C. Franklin}.}
  \bibinfo{year}{2023}\natexlab{}.
\newblock \showarticletitle{Large-area automatic detection of shoreline
  stranded marine debris using deep learning}.
\newblock \bibinfo{journal}{\emph{International Journal of Applied Earth
  Observation and Geoinformation}}  \bibinfo{volume}{124}
  (\bibinfo{year}{2023}), \bibinfo{pages}{103515}.
\newblock
\showISSN{1569-8432}
\urldef\tempurl%
\url{https://doi.org/10.1016/j.jag.2023.103515}
\showDOI{\tempurl}


\end{thebibliography}


\end{document}